\documentclass[journal]{IEEEtran}

\usepackage[style=ieee]{biblatex} 
\bibliography{example_bib.bib}    

\usepackage{amsmath}

\usepackage{url}
\usepackage{graphicx}  
\usepackage{float}  
\usepackage{CJK}
\usepackage[top=2cm, bottom=2cm, left=2cm, right=2cm]{geometry}
\usepackage{algorithm}
\usepackage{algorithmicx}
\usepackage{algpseudocode}
\usepackage{amsmath}
\usepackage{amssymb}
\usepackage{xspace}
 
\floatname{algorithm}{Algorithm}

\usepackage{color}
\usepackage{url}

\usepackage{booktabs}
\usepackage[hidelinks]{hyperref}

\usepackage{newpxtext}

\ExplSyntaxOn

\NewDocumentCommand { \ie } { }
  {
    \textit{i.e.}
    \peek_meaning_ignore_spaces:NTF .
      { \skip_horizontal:n { -.3ex } \use_none:n }
      {
        \peek_meaning_ignore_spaces:NF ,
          { \skip_horizontal:n { -.3ex } }
      }
  }
\NewDocumentCommand { \eg } { }
  {
    \textit{e.g.}
    \peek_meaning_ignore_spaces:NTF .
      { \skip_horizontal:n { -.3ex } \use_none:n }
      {
        \peek_meaning_ignore_spaces:NF ,
          { \skip_horizontal:n { -.3ex } }
      }
  }

\NewDocumentCommand { \etc } { }
  {
    etc.
    \peek_meaning_ignore_spaces:NT .
      { \use_none:n }
  }

\NewDocumentCommand { \etal } { }
  {
    et\ al.\
    \peek_meaning_ignore_spaces:NT .
      { \use_none:n }
  }

\NewDocumentCommand { \cad } { }
  {
    \textit{c-à-d.}
    \peek_meaning_ignore_spaces:NTF .
      { \skip_horizontal:n { -.3ex } \use_none:n }
      {
        \peek_meaning_ignore_spaces:NF ,
          { \skip_horizontal:n { -.3ex } }
      }
  }
\ExplSyntaxOff

\begin{document}

\title{Information Theory Inspired Pattern Analysis for Time-Series IoT Data}

\DeclareRobustCommand*{\IEEEauthorrefmark}[1]{%
  \raisebox{0pt}[0pt][0pt]{\textsuperscript{\footnotesize #1}}%
}


  \author{Yushan Huang,
      Yuchen Zhao,
      Alexander Capstick,
      Francesca Palermo,\\
      Hamed Haddadi,
      and Payam Barnaghi

  \thanks{Yushan Huang is with the Department of Computing, Imperial College London, and Care Research and Technology Centre, The UK Dementia  Research Institute, London, UK (e-mail: yushan.huang21@imperial.ac.uk).}
  \thanks{Yuchen Zhao is with the Department of Computer Science, University of York, York, UK (e-mail: yuchen.zhao@york.ac.uk).}%
  \thanks{Alexander Capstcik, Francesca Paleromo, and Payam Barnaghi are with the Department of Brain Sciences, Imperial College London, and Care Research and Technology Centre, The UK Dementia  Research Institute, London, UK (e-mail: {alexander.capstick19, f.palermo, p.barnaghi}@imperial.ac.uk).}
  \thanks{Hamed Haddadi is with the Department of Computing, Imperial College London, UK (e-mail: h.haddadi@imperial.ac.uk).}
  \thanks{Payam Barnaghi is also with the Great Ormond Street Institute of Child Health, University College London.}}

\maketitle

\begin{abstract}
Current methods for pattern analysis in time series mainly rely on statistical features or probabilistic learning and inference methods to identify patterns and trends in the data. Such methods do not generalize well when applied to multivariate, multi-source, state-varying, and noisy time-series data. To address these issues, we propose a highly generalizable method that uses information theory-based features to identify and learn from patterns in multivariate time-series data. To demonstrate the proposed approach, we analyze pattern changes in human activity data. For applications with stochastic state transitions, features are developed based on Shannon's entropy of Markov chains, entropy rates of Markov chains, entropy production of Markov chains, and von Neumann entropy of Markov chains. For applications where state modeling is not applicable, we utilize five entropy variants, including approximate entropy, increment entropy, dispersion entropy, phase entropy, and slope entropy. The results show the proposed information theory-based features improve the recall rate, F1 score, and accuracy on average by up to 23.01\% compared with the baseline models and a simpler model structure, with an average reduction of 18.75 times in the number of model parameters.

\end{abstract}


\section{Introduction}\label{Introduction}
\label{sections:introduction}


\IEEEPARstart{W}ith the development of small-scale and low-cost network-connected devices, large volumes of data is generated \cite{jobanputra2019human}. In particular, the Internet of Things (IoT) provides us with an unprecedented ability to capture real-world information. By integrating the real world with the digital world, IoT enables us to analyze and mine useful information based on collected data. These technologies have been widely applied across several fields such as healthcare \cite{selvaraj2020challenges}.

Time-series data is critical in the real world, as it contains key information on relationships from a temporal perspective. Analyzing time-series data facilitates the development of effective methods for observing the raw data and also allows us to understand relationships within the data. It also enables us to uncover the various patterns that exist in the data, determine the relationships between these patterns, analyze the trends, and make predictions. Unfortunately, the analysis of time-series data is very challenging, as such data (\eg, human activity data) is often multivariate \cite{zhang2019deep}, multi-source \cite{piccialli2021artificial}, rapidly state-varying \cite{tao2018novel}, and noisy \cite{both2021deepmod}, which is difficult to mine the potential information and can be easily affected by noise.

There are several well-established methods for pattern and trend analysis applied to time-series data \cite{zhu2017feature}. These methods can be classified into four categories based on their data mining approaches: statistical methods, statistical and probabilistic learning and inference methods, deep neural networks, and information theory-driven techniques. However, these methods have been limited in their applicability to multivariate, multi-source, rapidly state-varying, and noisy time-series data. Recently, deep neural network (DNN) models have attracted a great deal of attention. Such models can learn spatio-temporal properties of data, extract features automatically, and analyze patterns to predict outcomes or changes, such as state transitions\cite{portugal2019predicting}. Although deep neural network models can be effective in analyzing complex datasets, these models and the features they extract are often difficult to be interpreted. Interpretable features such as the features extracted by information theory-driven techniques can make a learning model surpass the performance of deep neural network models \cite{reyna2019early}, while also improving our ability to explain the inference process of machine learning models.

Our previous works include the Blocks of Eigenvalues algorithm for time series segmentation \cite{gonzalez2018beats} as a method to represent time-series data, a pattern representation method based on mutual information and entropy \cite{rezvani2019new}, and preliminary experiments and analysis of three Markov chain-based entropy features via heat maps \cite{huang2022using}. These studies highlight the potential of entropy features in analyzing time-series data. However, these works do not present a complete pipeline for analyzing time series data and do not validate the results of the methods by machine learning and deep learning models. These works have demonstrated the potential of using entropy when handling data that is multivariate, multi-source, rapidly state-varying, and noisy. Thus, inspired by information theory and entropy, in this paper we propose a pipeline to extract interpretable features in multivariate time-series data, which will enhance the performance of machine learning and deep learning models. The primary contributions of this paper are as follows:

(1) We introduce different entropy-based methods to derive engineered features from time-series data. We then propose a pipeline for extracting interpretable, higher-level features that are highly generalizable and applicable to processing multivariate, multi-source, rapidly state-varying and noisy time-series data.

(2) We apply our information theory-based models to one human activity dataset (from a clinical study for remote healthcare monitoring) and two publicly available datasets (Gait in Aging and Disease Database \cite{goldberger2000physiobank}, and PTB Diagnostic ECG Database \cite{bousseljot1995nutzung}) to demonstrate the applicability of this approach in different settings and for different applications.

(3) We evaluate the effectiveness of the extracted features using different models such as logistic regression, Support Vector Machines (SVM), Multi-Layer Perception (MLP), and Long Short Term Memory (LSTM) neural networks. Our experimental results show that, for the three different types of datasets, compared to the baseline methods, the information theory-based features can significantly improve the accuracy, recall, and F1 scores of the models by an average of 10\%-25\%.


In conclusion, we present a general pipeline for processing multivariate, multi-source, rapidly changing, and noisy time-series data. Our approach provides a comprehensive description of the creation, selection, and modeling of entropy features, offering a new perspective for analyzing this type of data. We also evaluate the effectiveness of our information theory-based pipeline using various datasets, showcasing its versatility and generalizability. Our approach has the potential to enhance the performance of machine learning models for time-series data analysis, making it a useful tool for real-world applications.

The remainder of this paper is organized as follows. In Section \ref{RelatedWorks}, we review the state-of-the-art works in pattern analysis for time-series data. In Section \ref{Datasets}, we introduce and analyze the original data from three datasets, which are multivariate, multi-source, rapidly state-varying and noisy. In Section \ref{Methodology}, to process this type of time-series data, we provide a technical description of the entropy techniques and their variants in detail to mine the potential information of the time-series data. In Section \ref{Modelling}, we demonstrate the evaluation results of the three datasets on different machine learning and deep learning models. Finally, in Section \ref{Conclusion}, we conclude our studies and discuss future work.

The source code, constructed models and links to the public datasets are made available via self-explanatory code with mark-up on a GitHub repository \cite{Yushan2023}.



\section{Related Works}\label{RelatedWorks}

There are four main approaches to mining information from time-series data: statistical methods, statistical and probabilistic learning and inference methods, deep neural network models, and information theory-driven techniques.

Classical statistical methods primarily focus on feature selection rather than data mining. However, with the increase in the amount and complexity of data, it becomes challenging to apply classical statistical theory-based techniques as they assume that the data is statistically uncorrelated. These techniques tend to perform poorly when applied to multivariate, multi-source, rapidly changing, or noisy time-series data \cite{ding2012survey}.

DNN-based techniques are popular for mining information from time-series data due to their ability to extract features and yield optimal results for large datasets. Feature extraction methods such as convolutional neural networks (CNN) and long short-term memory (LSTM) are typically used in the design of the DNN structure. In recent years, researchers have continuously carried out innovative research on the basis of these representative methods. For example, Xia \etal combined CNN and LSTM to create an eight-layer CNN-LSTM model that considers both spatial and temporal embedded information of the original data \cite{xia2020lstm}. Singh \etal added a self-attention mechanism to CNN-LSTM for better performance \cite{singh2020deep}. Despite the convenience of feature extraction using DNNs for time-series data, understanding and interpreting the extracted features is still a significant challenge due to the "black box" nature of DNNs. Furthermore, DNNs can only automatically extract simple features and not more complex features.

To mine features that are both interpretable and more complex from time-series data, some researchers have begun to develop research from the perspective of information theory. Shannon first proposed the concept of entropy, to measure the uncertainty of information, establishing the scientific theoretical basis of modern information theory \cite{shannon1948mathematical}. Based on Shannon's entropy, several entropy variants such as spectral entropy \cite{powell1979spectral} and sample entropy \cite{richman2004sample} have been proposed. Nurwulan \etal compared traditional features with multi-scale entropy (MSE) features extracted from 3-axis acceleration data and showed that MSE outperformed traditional features in KNN and random forest (RF) classification \cite{nurwulan2020multiscale}. Bao \etal extracted frequency-domain entropy features from original acceleration data, which were combined with mean, energy, and correlation of the original data as inputs to build a model with ideal results \cite{bao2004activity}. While the above entropy-based methods offer new avenues for time-series data analysis, they also have certain limitations. Many existing studies only utilize a single entropy feature or use entropy features as supplementary to traditional features. Furthermore, these methods are task-specific and do not form a comprehensive pipeline based on entropy methods. Additionally, there is a lack of a clear explanation for the selection and calculation of entropy features. 


A similar study to this paper is Howedi \etal's entropy measurement model \cite{howedi2019exploring}, which uses approximate entropy (ApEn), sample entropy (SampEn), and fuzzy entropy (FuzzyEn) to detect visitors in a home environment. However, this study does not select entropy features based on the data characteristics, such as Markovian systems and stochastic state transitions, and does not provide a justification for the selection of entropy features.

\begin{figure}[t]
\vskip 0.2in
\begin{center}
\centerline{\includegraphics[width=0.35\textwidth]{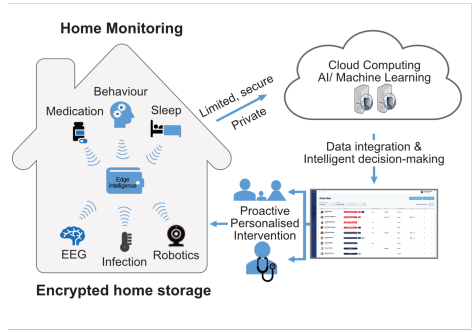}}
\caption{An overview of the healthcare monitoring IoT platform.}
\label{fig:network}
\end{center}
\vskip -0.2in
\end{figure}

\begin{table*}[t]
	\centering
	\caption{All IoT devices used in the Minder platform}
	\label{tab:Iot platform}
	\begin{tabular}{cccc} 
		\hline
		Digital Marker & Monitoring Device & Frequency \\
		\hline
		Human activity & Passive infrared sensors & Triggered by movement\\
		Home device usage & Smart plugs & Triggered by device use\\
		Body temperature & Smart  temporal  thermometers & Twice daily or continuous\\
		Blood pressure and heart rate & Wearable devices & Twice daily\\
		Weight and heart rate & Smart scale & Once a day\\
		Respiratory and heart rate during sleep & Sleep mat & Once a minute\\
		Environmental light & Light sensors & Every 15 minutes\\
		Environmental temperature & Temperature sensors & Once an hour\\
		\hline
	\end{tabular}
\end{table*}

\begin{figure}[t]
\vskip 0.2in
\begin{center}
\centerline{\includegraphics[width=0.35\textwidth]{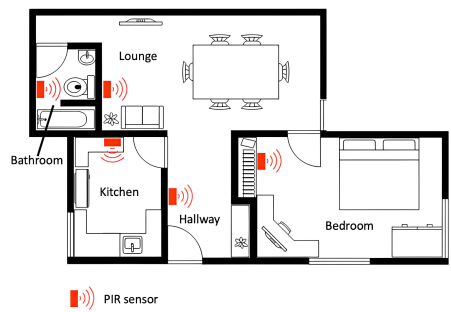}}
\caption{An example of PIR sensor installation in the study.}
\label{fig:AgeAndSensor}
\end{center}
\vskip -0.2in
\end{figure}

\begin{figure}[t]
\vskip 0.2in
\begin{center}
\centerline{\includegraphics[width=0.4\textwidth]{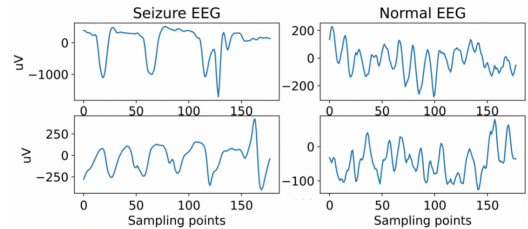}}
\caption{Visualisation of ESRD. The x-axis represents sampling points, and the y-axis represents EEG signals $(\mu V)$.}
\label{fig:GADD}
\end{center}
\vskip -0.2in
\end{figure}

\begin{figure}[t]
\vskip 0.2in
\begin{center}
\centerline{\includegraphics[width=0.4\textwidth]{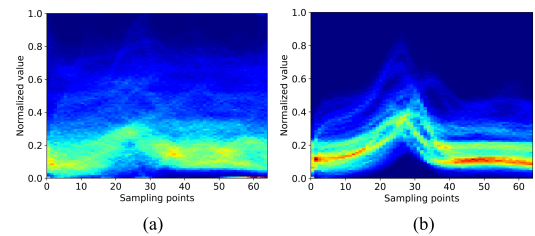}}
\caption{The histogram color maps for PTBDB marked as abnormal (a) and normal (b). The x-axis represents sampling points, and the y-axis represents the normalized value of the heartbeat.}
\label{fig:PTBDB}
\end{center}
\vskip -0.2in
\end{figure}

\section{Datasets}\label{Datasets}

In this paper, we apply our information theory-based pipeline to three datasets, one human activity data collected from the in-home healthcare monitoring IoT platform of our ongoing Minder study, as well as two publicly available EEG signal datasets, providing information on epileptic seizure and heart disease, respectively.

\subsection{Minder Dataset}

We have developed an in-home healthcare monitoring IoT platform (illustrated in Fig.~\ref{fig:network}), called Minder, to support people living with dementia (PLWD) \cite{enshaeifar2018health}. The Minder platform collects various digital markers, including activity data, home device usage, and clinical information. It comprises four main parts: 1) device-independent sensors installed in participants' homes to collect original data, 2) a back-end system with cloud infrastructure, storage, and analysis tools to analyze the data and install machine learning algorithms, 3) a user interface presenting clinical and environmental information, as well as alerts generated by the system, and 4) clinical intervention involving healthcare professionals using the system/alerts to communicate with participants and caregivers to address their medical needs. 

The Minder study protocol received ethical approval from the London-Surrey Borders Research Ethics Committee and South West London Ethics Committee (see \href{https://www.hra.nhs.uk/about-us/committees-and-services/res-and-recs/search-research-ethics-committees/london-surrey-borders/}{link here}) and we obtained informed written consent from all study participants. 

\begin{figure*}[t]
\vskip 0.2in
\begin{center}
\centerline{\includegraphics[width=0.8\textwidth]{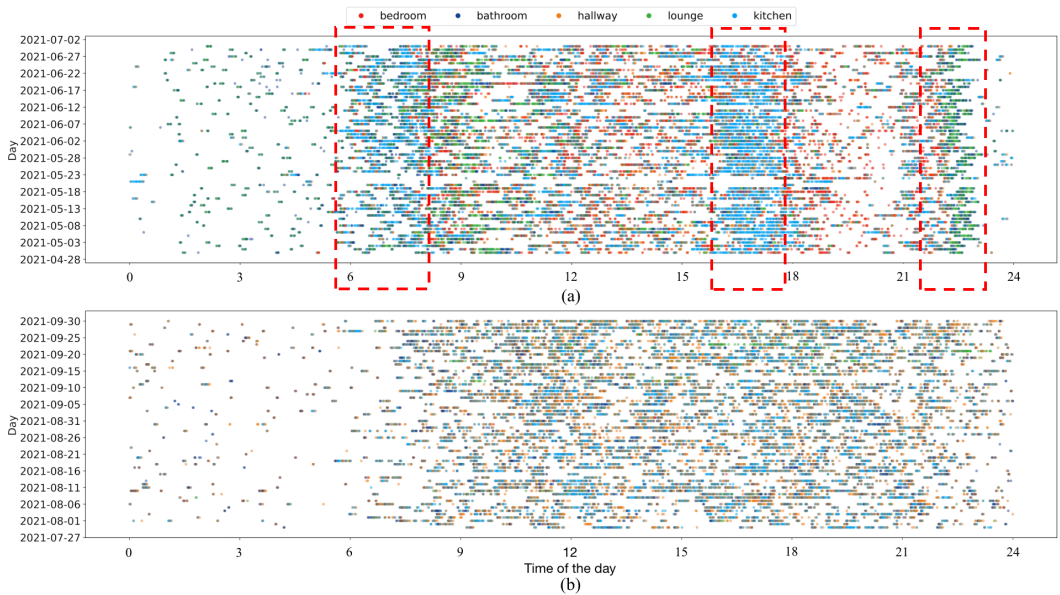}}
\caption{An example of a PLWD with clear routine activities (a), and another PLWD with fewer routine activities (b). The participant with more routine activities tends to have a more consistent daily activity pattern at the same time each day, as shown in the red boxes. The x-axis shows the time of the day, the y-axis shows different days, and the different colors represent different locations in the house.}
\label{fig:OriginalData}
\end{center}
\vskip -0.2in
\end{figure*}

The dataset is labeled by our monitoring team in response to alerts generated on the Minder platform, which operates 24/7. These alerts are verified with the people living with dementia (PLWD) or their caregivers, and provide information on potential healthcare-related events such as falls, abnormal motor function behavior, hospital admissions, Urinary Tract Infections, anxiety and depression, agitation, confusion, and disturbed sleep patterns. Participants who have experienced such events will have labeled data for that adverse health event.

In this study, we focus on the activity data of Minder only. This includes 3762 person-weeks of data collected between December 2020 and March 2022. The mean age of participants is 79. All of the data presented here has been anonymized. 

Activity data in the Minder platform is collected using PIR sensors installed in various locations, including the kitchen, bathroom, bedroom, lounge, and hallway, as shown in Fig.~\ref{fig:AgeAndSensor}. The PIR sensor logs an event with seconds precision and a 30-second delay when a person passes by. The recorded data is time-series data, showing the household's life patterns over time. We can identify clear differences in behaviors by visualizing the raw data, as shown in Fig.~\ref{fig:OriginalData}, which compares the routine activities of two PLWDs.

\subsection{Epileptic Seizure Recognition Dataset}

The ESRD (Epileptic Seizure Recognition Dataset) contains 11,500 time-series EEG signal data samples from 500 subjects and is used to study EEG signal changes during seizures \cite{andrzejak2001indications}. Each sample consists of 23 segments containing 178 data points over a one-second interval. The UCI preprocessed the original dataset and randomly rearranged the segments to form the 11,500 time-series EEG signal samples from 500 subjects. The dataset includes five different health conditions, including one related to epileptic seizures, and four normal conditions where the subjects do not show symptoms of epilepsy. However, many researchers choose to perform binary classification to distinguish between class 1 (representing epileptic seizures) and other classes. Our goal is also to distinguish between healthy participants and those with epileptic seizures.

\subsection{PTB Diagnostic ECG Database}

The PTB Diagnostic ECG Database (PTBDB) is a collection of 549 records from 290 subjects (209 male, and 81 female) \cite{bousseljot1995nutzung, goldberger2000physiobank}. The age range of participants is 17 to 87 years old, with an average age of 57.2. The sampling frequency is 125Hz. The Diagnostic class includes myocardial infarction, cardiomyopathy/heart failure, bundle branch block, dysrhythmia, myocardial hypertrophy, valvular heart disease, myocarditis, miscellaneous, and healthy controls. In this study, we extract heartbeat signals and only use ECG lead 2 \cite{kachuee2018ecg}. We focus on the myocardial infarction and healthy control categories, with a total of 14552 samples in the dataset. The histogram color maps for the PTB data marked as abnormal and normal are shown in Fig.~\ref{fig:PTBDB}.

\section{Methodology}\label{Methodology}

The pipeline proposed in this paper is mainly composed of three parts: data preprocessing, feature construction, and modeling.

The data preprocessing phase includes missing value processing, data resampling, and label encoding. The missing values are forward-filled with the last valid value, then backfilled with the next valid value. Data resampling is determined by the characteristics of the data as well as the requirements of the target. For example, if a dataset has a low sample size, but narrowing the time window has little impact on the target results, then resampling will be performed to expand the dataset. 

In the modeling stage, we use classical machine learning and deep learning models such as Logistic Regression (LR), Support Vector Machine (SVM), Multilayer perceptron (MLP), Convolutional neural network (CNN), and Long Short-term Memory (LSTM). 

The following introduces the feature construction stage, including the entropy and entropy variants used in this study, and the feature selection methods.

\subsection{Entropy and Entropy Variants}
\subsubsection{Shannon's Entropy of a Markov chain}
Assuming that a certain human activity (\eg, a sequence of locations) can form a Markov chain, then we can regard the occurrence of these activities as random events, and measure the extent of occurrence of these random events. We apply Shannon's entropy of a Markov chain to represent pattern changes in human activity data. Suppose that there are $n$ locations $X={x_1, x_2, ..., x_n}$ in a participant's activity, then the Shannon's entropy of a Markov chain $H(x)$ can be described as:

\begin{equation}
H(X)=-\sum_{i=1}^n P\left(x_i\right) \log P\left(x_i\right)
\end{equation}

In which $P(X_i)$ is the probability of activity $x_i$. When the frequency of a participant's  activity changes, $H(x)$ will change accordingly to represent the change in activity pattern.

\begin{figure*}[t]
\vskip 0.2in
\begin{center}
\centerline{\includegraphics[width=0.8\textwidth]{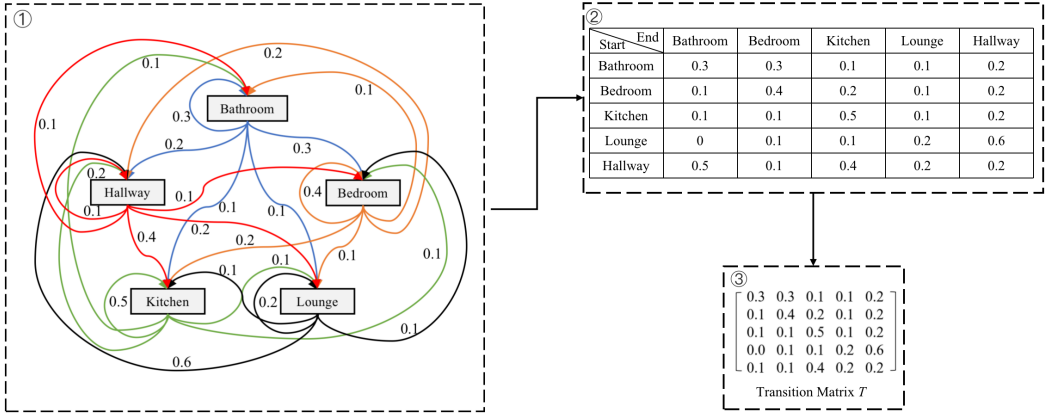}}

\caption{An example of the Entropy rate of a Markov chain. In 
\protect\textcircled{1}, the rectangular boxes represent the locations (states) in the Markov chain, and the arrows represent the routes between locations in the house. Different colors represent different start locations (blue: bathroom, orange: bedroom, green: kitchen, black: lounge, and red: hallway). The numbers next to the lines represent route probabilities which correspond to the table \protect\textcircled{2} and Transition Matrix $T$ \protect\textcircled{3}.}
\label{fig:EntropyRate}
\end{center}
\vskip -0.2in
\end{figure*}

\subsubsection{Entropy Rate of a Markov Chain}
Shannon's entropy of a Markov chain does not link the activities in a Markov chain together, but only treats each activity as a separate individual. However, if we utilize the first-order Markov chain to profile the human activities and collect these activities together, we can get the corresponding transitions, where the current activity event of a participant is only dependent on the preceding activity event \cite{lampard1968stochastic}. Suppose that $X=\{x_1, x_2, ..., x_n\}$ represents $n$ states in a Markov chain. Let $x_i, x_j \in X$, represent the previous state and the current state, respectively. Then the probability $P_{i j}$ of the route from $x_i$ to $x_j$ can be represented as:

\begin{equation}
P_{i j}=P\left(x_j \mid x_i\right)
\end{equation}

Where $x_i$ and $x_j$ $\in X$. Suppose that there are $n$ states in a Markov chain, then the Markov chain can be represented as $n \times n $ matrix ${P_{ij}}_{i,j \in X}$, called Transition Matrix $T$, an example is shown in Fig.~\ref{fig:EntropyRate}. From Markov chains, stationary distributions $\pi$ can be calculated, which represent:

\begin{equation}
\pi=\pi T
\end{equation}

In which, $\pi$ is an n-dimension vector associated with a Markov chain with $n$ states. Using this, the entropy rate of a Markov chain can be expressed as \cite{enshaeifar2018health}:

\begin{equation}
\xi=-\sum_{ij}^n \pi_i P_{i j} \log P_{i j}
\end{equation}

In which, $\pi_i$ is the probability in the stationary distribution associated with activity $x_i \in X$ in a Markov chain with the stationary distribution. When calculating the entropy rate of a Markov chain, there are two time-windows that need to be set, one time-window is used to calculate $P_{i j}$ for target time-series data, and the other is used to calculate $\pi_i$ to represent the characteristics of time-series data. The time window to calculate $P_{i j}$ is set by the mission objective. And it has to be noted that the time window to calculate the stationary distribution $\pi_i$ is important, as it should reflect the stationary pattern of the participant. For example, participants' routines may be affected by the seasons, then we need to avoid the possible effects of the seasons when setting up the time windows to calculate the stationary distribution, such as setting the time windows to override the seasonal variations. The complete procedure for calculating the Entropy Rate of a Markov Chain is shown in the Algorithm \ref{alg::EntropyRate}.

\begin{algorithm}[t]  
  \caption{Entropy rate of a Markov chain}  
  \label{alg::EntropyRate}  
  \begin{algorithmic}[1]  
    \State Define: $S=\{s_1, s_2, ..., s_L\}$ is a Markov chain trajectory, where $L$ is the length of the trajectory, and $s \in X, X=\{x_1, x_2, ..., x_n\}$, $n$ is the number of states in the Markov chain. $TW_1$ is the time window required for the stationary distribution, $TW_1 <= L$. $TW_2$ is the time window required for the target task, $TW_2 <= L$. $P_{i j}$ is the probability from state $x_i$ to state $x_j$. $SP$ is the start point;
    \Require  
      Markov chain trajectory $S$;  
    \Ensure  
      Entropy rate $\xi$ of the Markov chain; 
    \State Set $TW_1$ and $TW_2$;
    \Function {StationaryDistribution}{$S, TW_1$}
        \State $S_{TW_1} = S[0: TW_1]$;
        \State $P_{i j}^{\prime}=P\left(l_b^{\prime}=x_j^{\prime} \mid l_a^{\prime}=x_i^{\prime}\right)$, where $l_a^{\prime},l_b^{\prime} \in X$, represent the previous state and the current state, $a^{\prime} \in [2, TW_1], b^{\prime} \in [1, TW_1-1], x_i^{\prime} \in X, x_j^{\prime} \in X$;
        \State $\pi=\pi T$;
        \State \Return{$\pi$};
    \EndFunction
    
    \Function {EntropyRate}{$\pi, S, TW_2$}

        \For {$SP = 0$; $SP + TW_2 <= L$; $SP=SP+TW_2$}
        \State $S_{TW_2} = S[SP: TW_2]$;
        \State $P_{i j}=P\left(l_b=x_j \mid l_a=x_i\right)$, where $l_a,l_b \in X$, represent the previous state and the current state, $a \in [2, TW_2], b \in [1, TW_2-1], x_i \in X, x_j \in X$;
        \State $\xi_m=-\sum_{x_i, x_j \in X}^n \pi_i P_{i j} \log P_{i j}$;
        \EndFor
        \State \Return{$\xi=\{\xi_1, \xi_2, ..., \xi_m\}$};

    \EndFunction
    
  \end{algorithmic}  
\end{algorithm}

\subsubsection{Entropy Production of a Markov Chain}

Entropy Production (EP) is a description of the diverse non-equilibrium principle \cite{dewar2003information}, which is intended to describe physical processes. Physical processes can be described by stochastic processes, such as Markov chains and diffusion processes. The Markov chains generated by human activity data can be regarded as a stochastic process \cite{abdelgawwad20193d}. Therefore, we can apply EP to Markov chains to describe the pattern changes.

EP can be estimated by ML models such as the Neural
Estimator for Entropy Production (NEEP), which can estimate EP of Markovian systems \cite{kim2020learning}. Given a Markov chain trajectory $S = \{s_1, s_2, . . . , s_L\}$ and a
function $h_\theta$ acting over previous state $s_t$ and the current state $s_{t+1}$ in the Markov chain, where $\theta$ denotes the trainable neural network parameters, then the output of NEEP can be defined as \cite{kim2020learning}:

\begin{equation}
\hat{J}(\theta)=\sum_{t \in L}\left[\Delta S_\theta\left(s_t, s_{t+1}\right)-e^{-\Delta S_\theta\left(s_t, s_{t+1}\right)}\right]
\end{equation}

Where $\Delta S_\theta$ is:

\begin{equation}
\Delta S_\theta\left(s_t, s_{t+1}\right) \equiv h_\theta\left(s_t, s_{t+1}\right)-h_\theta\left(s_{t+1}, s_t\right)
\end{equation}

\begin{figure}[t]
\vskip 0.2in
\begin{center}
\centerline{\includegraphics[width=0.45\textwidth]{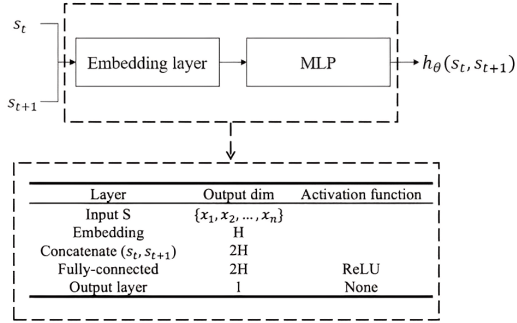}}
\caption{The model structure of NEEP of Markovian systems. $H$ is the size of the embedding dimension.}
\label{fig:NEEP}
\end{center}
\vskip -0.2in
\end{figure}

The procedure for training NEEP is shown in  Algorithm \ref{alg::NEEPtraining} and the model structure of NEEP is shown in Fig.~\ref{fig:NEEP}. In NEEP, an embedding layer is used to transform the discrete state into a trainable continuous vector \cite{kim2020learning}, then the embedded data is input into a hidden MLP layer. It has to be noted that, the length of the time series data is very important when training NEEP, as we need to ensure that the data for this period of time is sufficient for training and can reflect the participant's characteristics.

\begin{algorithm}[t]  
  \caption{Training process of NEEP}  
  \label{alg::NEEPtraining}  
  \begin{algorithmic}[1]  
    \State Define: $S=\{s_1, s_2, ..., s_L\}$ is a Markov chain trajectory, where $L$ is the length of the trajectory, and $s \in X, X=\{x_1, x_2, ..., x_n\}$, $n$ is the number of states in the Markov chain.
    \Require  
      Markov chain trajectory $S$; 
    \Ensure  
      The values calculated by the loss function $\hat{J}(\theta)$; 
    \Loop
        \State Embedding layer;
        \State Objective function 
            \begin{equation}
            \hat{J}(\theta)=\sum_{t \in {L}}\left[\Delta S_\theta\left(s_t, s_{t+1}\right)-e^{-\Delta S_\theta\left(s_t, s_{t+1}\right)}\right]
            \end{equation}
        \State Compute gradients $\nabla_\theta \hat{J}(\theta)$;
        \State Update parameters $\theta$;
    \EndLoop
  \end{algorithmic}  
\end{algorithm}

\begin{figure*}[t]
\vskip 0.2in
\begin{center}
\centerline{\includegraphics[width=1.0\textwidth]{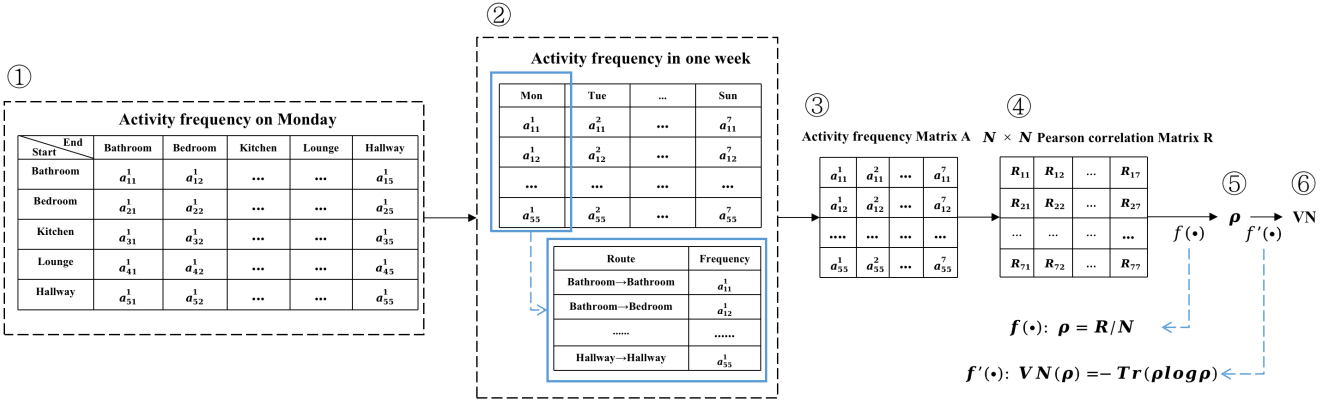}}
\caption{An example for the von Neumann entropy of a Markov chain. Suppose that there are five locations (states) in a Markov chain, and we plan to calculate the von Neumann entropy of one week. From the perspective of spatial, \protect\textcircled{1}: count the frequency $a_{i,j}^d$ of different routes in a Markov chain for each day of the week, in which $i$ and $j$ represent the previous location and the current location, respectively; \protect\textcircled{2}: aggregate weekly activity frequency; \protect\textcircled{3}: transfer the weekly activity frequency to the activity frequency matrix $A$; \protect\textcircled{4}: calculate the Pearson correlation Matrix $R$ between each day; \protect\textcircled{5}: calculate the density matrix $\rho$ by $f(\bullet)$; \protect\textcircled{6}: calculate the von Neumann entropy by $f^{\prime}(\bullet)$. And from the perspective of temporal, the only difference is changing the activity frequency to activity duration.}
\label{fig:VN}
\end{center}
\vskip -0.2in
\end{figure*}

\begin{algorithm}[t]  
  \caption{von Neumann entropy of a Markov chain}  
  \label{alg::VN}  
  \begin{algorithmic}[1] 
    \State Define: $S=\{s_1, s_2, ..., s_L\}$ is a Markov chain trajectory, where $L$ is the length of the trajectory, and $s \in X, X=\{x_1, x_2, ..., x_n\}$, $n$ is the number of states in the Markov chain. $VN$ is the von Neumann entropy of a Markov chain. $TW_2$ is the time window required for the target task, where $TW_2 <= L$. $SP$ is the start point;
    \Require  
      Markov chain trajectory $S$;  
    \Ensure  
      The $VN$; 
    \State Set $TW_2$;
    \For{$SP = 0$; $SP + TW_2 <= L$; $SP=SP+TW_2$}
        \State Calculate original matrix $A$ (e.g., activity frequency matrix);
        \State $N \times N$ Pearson correlation Matrix $R$ of $A$;
        \State Density operator $\rho \gets \rho = R/N$;
        \State von Neumann entropy $VN(\rho) \gets VN(\rho)=\operatorname{Tr}(\rho \log \rho)$, 
        $log\rho = \sum_{k=1}^{\infty}(-1)^{k+1} \frac{(B-I)^k}{k}$;
    \EndFor
\end{algorithmic}  
\end{algorithm}

\subsubsection{von Neumann Entropy of a Markov Chain}
The von Neumann entropy (VNE) quantifies the amount of information present in a system, which can be applied to time-series data to quantify the fluctuation and the correlation of the data \cite{bengtsson2017geometry}. For a density operator $\rho$ with $N$ eigenvalues $\lambda_{1, \ldots, n}$, VN is defined as follows:

\begin{equation}
S(\boldsymbol{\rho})=-\operatorname{tr}(\boldsymbol{\rho} \log \boldsymbol{\rho})=-\sum_{j=1}^N \lambda_j \log \lambda_j
\end{equation}

We apply VNE to the human activity data with stochastic state transitions to reflect the pattern change of the data. The human activity data of a Markov chain can be analyzed by VN from spatial and temporal perspectives, for example, illustrated in Fig.~\ref{fig:VN}. One of the key points to calculate VN is to obtain the density operator $\rho$, which must satisfy (i) be Hermitian, (ii) have unit trace, and (iii) be positive semidefinite. Given $\boldsymbol{R} \in \mathbb{R}^N$, an N-dimension Pearson correlation matrix of the human activity data, then the density operator $\boldsymbol{\rho}$ can be defined as \cite{felippe2021neumann}:

\begin{equation}
\boldsymbol{\rho}=\boldsymbol{R} / N
\end{equation}

The density operator $\rho$, calculated by Eq. (9) satisfies all the requirements. However, it has to be noted that the density operator $\rho$, which is calculated from real IoT data, may be sparse and thus there may be anomalies in the calculation of $log\rho$ using standard classical mathematical methods. Therefore, we calculate $\log\rho$ by Mercator's Series. Suppose $B$ is a matrix and sufficiently close to the identity matrix $I$, and $\|B-I\|<1$, then a logarithm of $B$ can be computed by means of the following k-power series \cite{macduffee2012theory}:

\begin{equation}
\log (B)=\sum_{k=1}^{\infty}(-1)^{k+1} \frac{(B-I)^k}{k}
\end{equation}

This means, we can obtain $\log\rho$ by:
\begin{equation}
\log (\boldsymbol{\rho})=\sum_{k=1}^{\infty}(-1)^{k+1} \frac{(\boldsymbol{\rho}-I)^k}{k}
\end{equation}

Integrating Eq. (8), Eq. (9) and Eq. (10), the VN can be obtained. The complete procedure for calculating VNE is shown in the Algorithm \ref{alg::VN}.

\subsubsection{Approximate Entropy}
For Non-Markovian chain systems, Approximate Entropy (ApEn) can be used to quantify the complexity of the system. Given a time series dataset $\{u(i): 1 \leq i \leq N\}$ with $N$ samples, form the sequence in order to generate an m-dimension vector:

\begin{equation}
u^{\prime}(i)=[u(i), u(i+1), \ldots, u(i+m-1)] \quad i=1, N-m+1
\end{equation}

Define the distance between the vectors $u^{\prime}(i)$ and $u^{\prime}(j)$ to be the maximum of the differences between the corresponding elements of the two vectors:

\begin{equation}
d[u^{\prime}(i), u^{\prime}(j)]=\max _{k=0, m-1}[|u^{\prime}(i+k)-u^{\prime}(j+k)|]
\end{equation}

Given a threshold $p$, count the number of $d[u^{\prime}(i), u^{\prime}(j)] <= p$, denoted as $A_N^m(p)$, and calculate the ratio of $A_N^m(p)$ to $N-m+1$, denoted as $B_N^m(p)$:

\begin{equation}
B_l^m(p)=\frac{A_l^m(p)}{N-m+1}
\end{equation}

Calculate the average value of $B_N^m(p)$:

\begin{equation}
B^m(p)=\frac{1}{N-m+1} \sum_{N=1}^{N-m+1} B_N^m(p)
\end{equation}

Increase the dimension from $m$ to $m+1$, and repeat the above steps. For sequences of finite length, an estimate of the sample entropy can be obtained as \cite{chen2006comparison}:

\begin{equation}
\operatorname{ApEn}(m, r, N)=B^m(p)-B^{m+1}(p)
\end{equation}

\subsubsection{Increment Entropy}
The Incremental Entropy (IncrEn) algorithm is a method for calculating the entropy of a sequence of data points incrementally, rather than computing the entropy of the entire sequence all at once. Given a time series dataset $\{u(i): 1 \leq i \leq N\}$ with $N$ samples. Construct an increment time series $\{v(i), 1 \leq i \leq N-1]$ by $v(i)=x(i+1)-x(i)$ from $u(i)$. Hence, for a positive integer $m$, $N-m$ vectors of dimension $m$ are derived from an incremental time series. These vectors, denoted as $V(k) =[v(k), v(k+1), \ldots, v(k+m-1)], 1 \leq k \leq N-m$, represent contiguous segments of the time series. Each element in a vector $V(k)$ is mapped onto a word of two letters. The sign of each component is represented by $v^{\prime}_{k+j}=\operatorname{sgn}(v(k+j)), j=1 \cdots, m-1$, and the magnitude of each component in relation to the other components within the vector is represented by $q_{k+j}, j=1, \ldots, m-1$ for a quantifying resolution $r$. As a result, $N-m$ words, ${w_k, 1 \leq k \leq N-m}$, are generated. Each word, consisting of $2 \times m$ letters, can have $(2r+1)^m$ variations, depending on the values of $m$ and $r$. The frequency of occurrence of each unique word $w_n$ is defined as:

\begin{equation}
p\left(w_n\right)=\frac{Q\left(w_n\right)}{N-m}
\end{equation}

where $Q\left(w_n\right)$ signifies the count of the unique word $w_n$ within the $\left\{w_k\right\}$. The Increment Entropy (IncrEn) of order m (where $m$ is equal to or greater than 2) and resolution $R$ is defined as:
\begin{equation}
IncrEn(m)=-\sum_{n=1}^{(2 R+1)^m} p\left(w_n\right) \log p\left(w_n\right)
\end{equation}

\subsubsection{Dispersion Entropy}

Dispersion entropy (DE) can be used to describe the complexity of time series data. For time series with low regularity, DE can reflect the degree of disorder of the series \cite{rostaghi2016dispersion}. Given a time series dataset $\{u(i): 1 \leq i \leq N\}$ with $N$ samples. Map $u(i)$ to $y(i)$ between 0 and 1 by normal cumulative distribution function (NCDF):

\begin{equation}
y_j=\frac{1}{\sqrt{2 \pi} \sigma} \int_{-\infty}^{u_j} e^{-\left((t-\mu)^2 / 2 \sigma^2\right) d t}
\end{equation}

In which, the parameter $\mu$  is the expectation of $u(i)$, while the parameter $\sigma$  is its standard deviation. Map $y$ to the range of $[1, 2, ..., c]$, and obtain a new sequence $z_j^{(c)}$:

\begin{equation}
z_j^{(c)}=\operatorname{int}\left(c y_j+0.5\right)
\end{equation}

In which, $c$ is the number of categories, and $int$ is the
rounding function. Then construct the embedding vector $z_i^{(m, c)}$ by:

\begin{equation}
\begin{aligned}
z_i^{(m, c)}= & \left(z_i^{(c)}, z_{i+d}^{(c)}, \cdots, z_{i+(m-1) d}^{(c)}\right), &  \\
& \quad i=1,2, \cdots, N-(m-1) d 
\end{aligned}
\end{equation}

\begin{figure}[t]
\vskip 0.2in
\begin{center}
\centerline{\includegraphics[width=0.45\textwidth]{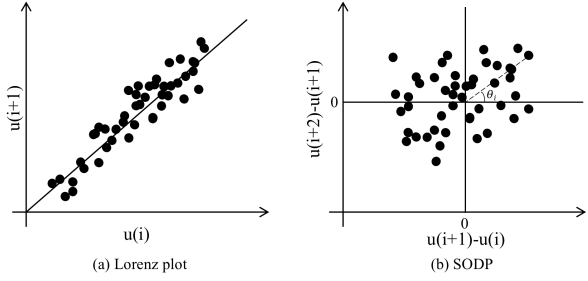}}
\caption{The phase space representation of an HRV signal.}
\label{fig:PhEn}
\end{center}
\vskip -0.2in
\end{figure}

In which, $m$ is the embedding dimension, $c$ is the number of class, $d$ is the time delay. Then each $z_j^{(m, c)}$ is mapped to dispersion pattern $\pi_{v_0 v_1 \cdots v_{m-1}}(v=1,2, \cdots, c)$, in which $z_i^{(c)}=v_0$, $z_{i+d}^{(c)}=v_1$, $...$, and $z_{i+(m-1) d}^{(c)}=v_{m-1}$. The number of possible dispersion of each $z_j^{(m, c)}$ is $c^m$.

Calculate the relative frequency for each potential dispersion pattern:

\begin{equation}
P\left(\pi_{v_0 v_1, \cdots, v_{m-1}}\right)=\frac{\operatorname{num}\left(\pi_{v_0 v_1, \cdots, v_{m-1}}\right)}{N-(m-1) d}
\end{equation}

Finally, based on Shannon's entropy, DE can be obtained by \cite{chakraborty2021automated}:

\begin{equation}
D E(u, m, c, d)= -\sum_{\pi=1}^{c^m} p\left(\pi_{v_0, \ldots, v_{m-1}}\right) \ln \left(p\left(\pi_{v_0, \ldots, v_{m-1}}\right)\right)
\end{equation}

\subsubsection{Phase Entropy}
Phase entropy (PhEn) is developed to detect the complexity of physiological signals. For example, given a time series dataset $\{u(i): 1 \leq i \leq N\}$ with $N$ samples, we can represent the data by the Lorenz plot, as Fig.~\ref{fig:PhEn} (a) shows. In the Poincaré plot, if we replace the sequence $u_i$ by $u_{i + 1} - u_i$, then we can get SODP plot, as Fig.~\ref{fig:PhEn} (b) shows. Specifically, from a given time series $u_i$, we can obtain $Y_i$ and $X_i$ by \cite{rohila2019phase}:

\begin{equation}
\begin{aligned}
& Y_i=u_{i+2}-u_{i+1} \\
& X_i=u_{i+1}-u_i
\end{aligned}
\end{equation}

Then compute the slope angle of each scatter point as shown in Fig (b).

\begin{equation}
\theta_i=\tan ^{-1} \frac{Y_i}{X_i}
\end{equation}

Then the probability distribution $p_i$ can be calculated by:

\begin{equation}
p_i = \frac{S_{\theta_i}}{\sum_{i=1}^k S_{\theta_i}}
\end{equation}

Finally, based on Shannon's entropy, the PhEn can be calculated as \cite{rohila2019phase}:

\begin{equation}
\operatorname{PhEn}=\frac{-1}{\log N} \sum_{i=1}^k p(i) \log p(i)
\end{equation}

\subsubsection{Slope Entropy}

Slope Entropy (SlopEn) is an algorithm to describe the complexity of a time series dataset, which is primarily based on transferring the original time series data to a series of single-threshold and symbolic patterns \cite{li2022particle,li2022novel}. Given a time series dataset $\{u(i): 1 \leq i \leq N\}$ with $N$ samples. Decompose $u$ into $j$ subsequences according to the embedded dimension $m$:

\begin{equation}
u_i^m=\left\{u_i, u_{i+1}, \cdots, u_{i+m-1}\right\}
\end{equation}

In which, $i = \{1,2,...,j\}$, $j=N-m+1$. Define two soft threshold parameters $\delta$ and $\gamma$ to calculate the symbolic patterns of $u_i^m$, where $0<\delta<\gamma$.

Define $d=u_{i+1}-u_i$, and compare $d$ with the two soft threshold parameters $\delta$ and $\gamma$, then five patterns can be obtained:

\begin{equation}
\begin{cases}\text { pattern }=2, & \gamma<d, \\ \text { pattern }=1, & \delta<d \leq \gamma, \\ \text { pattern }=0, & |d| \leq \delta, \\ \text { pattern }=-1, & -\gamma \leq d<-\delta, \\ \text { pattern }=-2, & d<-\gamma .\end{cases}
\end{equation}

Based on the five patterns, we can get $5^{m-1}$ sequence
combinations. The relative frequency $p_n$ of the
the combination can be calculated by the number of occurrences $f_n$ of each combination:

\begin{equation}
p_n=\frac{f_n}{j}, n=1,2, \cdots, 5^{m-1}
\end{equation}

Finally, SlopEn can be calculated based on the Shannon's entropy: 

\begin{equation}
\operatorname{SE}(m, \gamma, \delta)=-\sum_{n=1}^{5^{m-1}} p_n \ln p_n
\end{equation}

\begin{figure*}[t]
\vskip 0.2in
\begin{center}
\centerline{\includegraphics[width=0.8\textwidth]{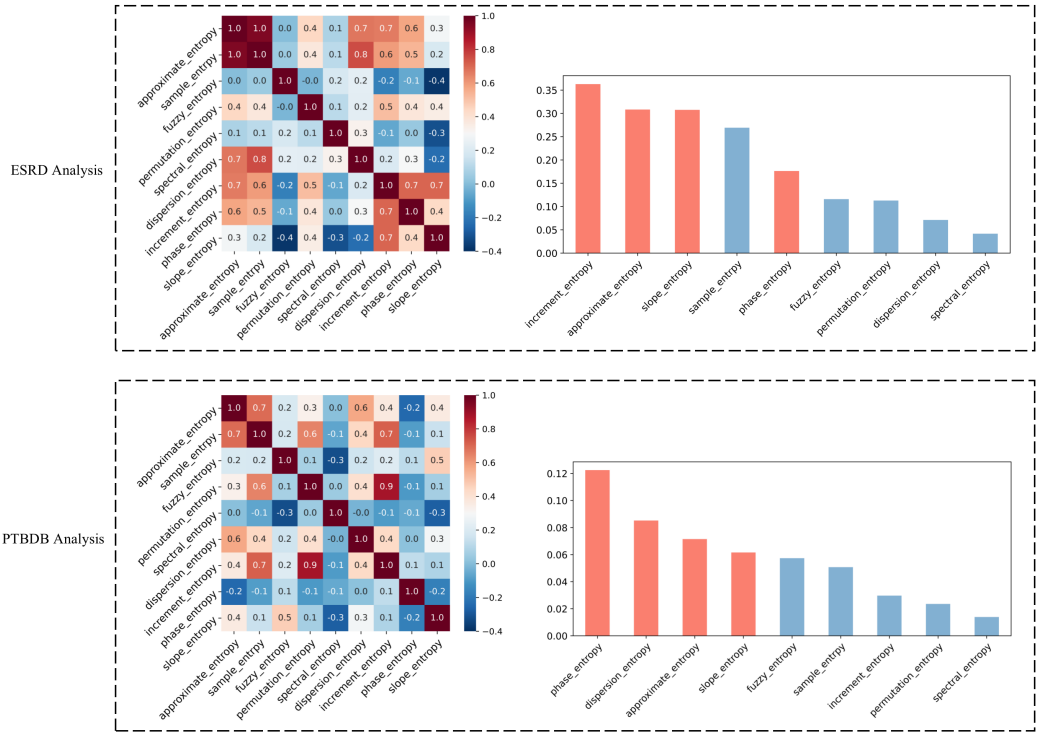}}
\caption{Feature selection of ESRD and PTBDB, including Pearson relationship matrices (the left) and mutual information (the right). The coordinates of the Pearson correlation matrices and the horizontal coordinates of the mutual information represent some common entropy measures. In mutual information features, the red represents the final selection of entropy features. For ESRD, according to the mutual information, although the most four important features are increment entropy, approximate entropy, slope entropy, and sample entropy, the Pearson correlation coefficient is too high between approximate entropy and sample entropy. Thus we finally select phase entropy to replace sample entropy.}
\label{fig:FeatureSelection}
\end{center}
\vskip -0.2in
\end{figure*}

\subsection{Feature Selection}
For feature selection, if the dataset is with stochastic state transitions and can be constructed as a Markov chain, we prioritize the entropy features associated with Markov chains, because linking the time-series data together to form Markov chains can potentially mine more information. For the dataset where state-space modeling is not applicable, we utilize mathematical statistics such as mutual information and the Pearson relationship matrix for filtering.

\subsubsection{Minder Database}
As Fig.~\ref{fig:OriginalData} shows, the original data of the Minder Database mainly includes the time and location where the infrared sensors were triggered, and the original data can be reconstructed into Markov chains to reflect the activity routes of the participants. Therefore, we prioritize the entropy features associated with Markov chains, including Shannon's entropy, the entropy rate of a Markov chain, the entropy production of a Markov chain, and the von Neumann entropy of a Markov chain (from the perspective of spatial and temporal). 

\subsubsection{ESRD and PTBDB}
As Fig.~\ref{fig:GADD} and Fig.~\ref{fig:PTBDB} show, the data from ESRD and PTBDB is collected by wearable sensors, and it is hard to generate Markov chains. Thus we apply mutual information and Pearson relationship matrices to ESRD and PTBDB to select approximate entropy features, as  Fig.~\ref{fig:FeatureSelection} shows. 

\section{Modeling and Results} \label{Modelling}

We utilize classical models to evaluate the entropy features, including Logistic Regression (LR), Support Vector Machine (SVM), Multilayer Perceptron (MLP), Convolutional neural network (CNN), and Long-short Term Memory (LSTM). 

\begin{figure*}[t]
\vskip 0.2in
\begin{center}
\centerline{\includegraphics[width=1\textwidth]{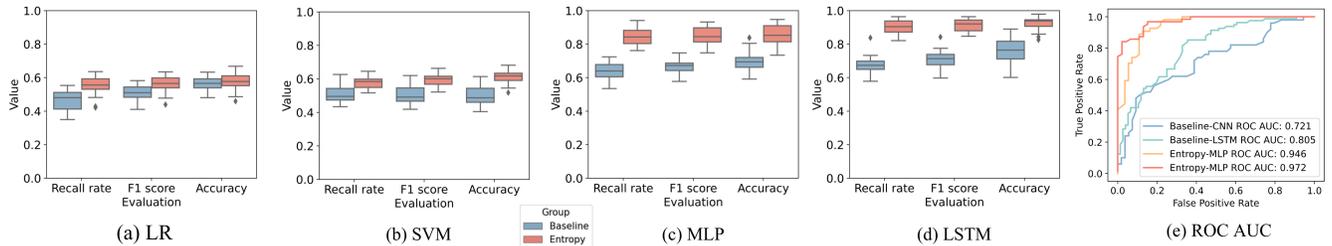}}
\caption{The evaluation results of the Minder database. (a), (b), (c), and (d) are the results of LR, SVM, MLP, and LSTM, respectively. The x-axis represents different evaluation methods.  And (e) is the ROC-AUC curves for MLP and LSTM models. From (a), (b), (c), and (d), we can find that, for four different models, compared with the baseline features, modeling by the entropy features can improve the recall rate, F1 score, accuracy, and AUC score. Especially for LSTM, the entropy features can improve the recall rate (90.29\%), F1 score (91.29\%), and accuracy (92.41\%) by about 23.01\%, 20.04\%, and 16.35\%. }
\label{fig:ResultMinder}
\end{center}
\vskip -0.2in
\end{figure*}

\begin{table}[t]
\centering
\caption{The average performance of the models for the Minder}
\label{tab:PerformanceMinder}
\begin{tabular}{@{}ccccc@{}}
\toprule
     &  Evaluation & Baseline & Entropy & Improvement \\ \midrule
LR   & Recall rate & 46.42±6.11\%  & 55.41±5.67\% & 8.79\%     \\
     & F1 score    & 50.76±4.53\%  & 56.59±4.67\% & 5.46\%     \\
     & Accuracy    & 56.52±4.13\%  & 58.03±4.70\% & 1.52\%      \\
SVM  & Recall rate & 51.11±5.24\%  & 57.84±3.35\% & 6.73\%      \\
     & F1 score    & 50.61±5.48\%  & 59.34±3.65\% & 8.73\%     \\
     & Accuracy    & 50.13±5.76\%  & 60.94±4.03\% & 10.81\%      \\
MLP  & Recall rate & 63.76±5.03\%  & 84.16±5.21\% & 20.40\%     \\
     & F1 score    & 66.59±3.71\%  & 84.97±4.97\% & 18.38\%     \\
     & Accuracy    & 70.15±5.85\%  & 85.88±5.51\% & 15.73\%     \\
LSTM & Recall rate & 67.28±4.99\%  & \textbf{90.29±4.41\%} & \textbf{23.01\%}     \\
     & F1 score    & 71.25±4.99\%  & \textbf{91.29±3.72\%} & \textbf{20.04\%}     \\
     & Accuracy    & 76.06±7.15\%  & \textbf{92.41±4.18\%} & \textbf{16.35\%}     \\ \midrule
Average & Recall rate & -  & - & \textbf{14.73}\%     \\
        & F1 score & -  & - & \textbf{13.15}\%     \\
        & Accuracy & -  & - & \textbf{11.10}\%     \\\bottomrule
\end{tabular}
\end{table}

\subsection{Minder Database}
We evaluate the performance of LR, SVM, MLP, and LSTM on the Minder database. Since our focus is on identifying whether a participant has had any non-healthy events, we use recall rate, F1 score, and accuracy as evaluation methods. Additionally, we consider the effect of sundowning and circadian rhythms in people living with dementia (PLWD) \cite{volicer2001sundowning} by dividing one day into two time periods: daytime (06:00 - 18:00) and night (18:00 - 24:00 and 00:00 - 6:00). The baseline features are average frequency of bathroom, bedroom, hallway, kitchen, and lounge in each week (daytime and night). The entropy features are Shannon's entropy of Markov chains, Entropy rate of Markov chains, EP of Markov chains, VNE of Markov chains (activity frequency), VNE of Markov chains (activity duration), and activity duration difference of Markov chains in each week (daytime and night). The output of the models is healthcare-related events (True or False).

\textbf{LR:} Model parameters of the baseline features: penalty = L2, solver = sag, class weight = balanced, random state = 10, test size = 0.3, repeat times = 30. Model parameters of the entropy features: penalty = L2, solver = sag, class weight = balanced, random state = 10, test size = 0.3, repeat times = 30.

\textbf{SVM:} Model parameters of the baseline features: kernel = linear, test size = 0.3, repeat times = 30. Model parameters of the entropy features: kernel = linear, test size = 0.3, repeat times = 30.

\textbf{MLP:} Model parameters of the baseline features: input layer ($10$), hidden layer ($10 \times 30, 30 \times 30$), output layer ($30 \times 1$), activation functions = ($tanh, tanh, sigmoid$), epochs = 3000, batch size = 256, learning rate = 0.15, criterion = Binary Cross-Entropy, optimizer = SGD, test size = 0.3, repeat times = 30. Model parameters of the entropy features: input layer ($1 \times 12$), hidden layer ($12 \times 50, 50 \times 50$), output layer ($50 \times 1$), activation functions = ($tanh, tanh, sigmoid$), epochs = 5000, batch size = 256, learning rate = 0.06, criterion = Binary Cross-Entropy, optimizer = SGD, test size = 0.3, repeat times = 30.

\textbf{LSTM:} Model parameters of the baseline features: input layer ($10$), hidden layer ($10 \times 30, 30 \times 30$), output layer ($30 \times 1$), activation functions = ($tanh, tanh, sigmoid$), epochs = 5000, batch size = 256, learning rate = 0.15, criterion = Binary Cross-Entropy, optimizer = SGD, timesteps = 3, test size = 0.3, repeat times = 30. Model parameters of the entropy features: input layer ($1 \times 12$), hidden layer ($12 \times 50, 50 \times 50$), output layer ($50 \times 1$), activation functions = ($tanh, tanh, sigmoid$), epochs = 5000, batch size = 256, learning rate = 0.4, criterion = Binary Cross-Entropy, optimizer = SGD, timesteps = 3, test size = 0.3, repeat times = 30.

The results of the Minder Database are shown in Fig.~\ref{fig:ResultMinder} and Table.~\ref{tab:PerformanceMinder}. We can find that, compared with the baseline features, modeling with the entropy features can improve the recall rate, F1 score, and accuracy on average by 14.03\%, 13.86\%, and 11.10\%. Especially for LSTM, compared with the model build using baseline features, the recall rate (90.29\%), F1 score (91.29\%), and Accuracy (92.41\%) are improved by 23.01\%, 20.04\%, and 16.35\%. 

\begin{figure*}[t]
\vskip 0.2in
\begin{center}
\centerline{\includegraphics[width=1\textwidth]{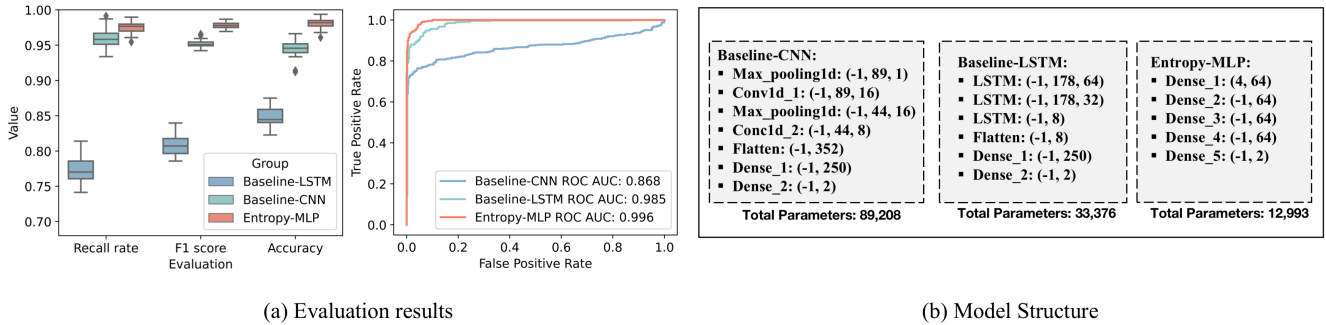}}
\caption{The results of ESRD. (a) is the evaluation results, including recall rate, F1 score, accuracy, and ROC-AUC. (b) is the comparison of the model structure between the baseline and entropy models. From (a), we can find that the recall rate, F1 score, and accuracy of the entropy-MLP models can be improved to up to 97.51\%, 97.80\%, and 98.10\%. As we used pre-processed data with fewer data noise, the AUC-ROC performances of all the models are ideal. From (b), we can find that the entropy-MLP model is reduced by 5.86 times and 1.59 times in terms of total parameters.}
\label{fig:ResultGADD}
\end{center}
\vskip -0.2in
\end{figure*}

\begin{table}[t]
\centering
\caption{Comparison of ESRD classification results}
\label{tab:PerformanceGADD}
\begin{tabular}{@{}ccccc@{}}
\toprule
           &  Recall rate & F1 score & Accuracy \\ \midrule
Baseline-CNN & 77.27±1.99\%  & 80.91±1.54\% & 84.93±1.38\%     \\
Baseline-LSTM& 96.01±1.47\%  & 95.21±0.66\% & 94.46±1.11\%      \\
Entropy-MLP   & \textbf{97.51±0.81}\%  & \textbf{97.80±0.45\%} & \textbf{98.10±0.72\%}\\\midrule
Avg Improvement   & \textbf{10.87\%}  & \textbf{9.74\%} & \textbf{8.41\%}\\\midrule
\end{tabular}
\end{table}

\subsection{Epileptic Seizure Recognition Dataset}

We aim to differentiate between the normal participants and those with epileptic seizures. The baseline models are LSTM and CNN with complete data. The entropy model is MLP with IncrEn, ApEn, SlopEn, and PhEn. The output of the models is Participants with epileptic seizures (True or False).

\textbf{The Baseline-CNN:} Max pooling-1d layer 1 ($-1 \times 89 \times 1$), Conv-1d layer 1 ($-1 \times 89 \times 16$), Max pooling-1d layer 2 ($-1 \times 44 \times 16$), Conv-1d layer 2 ($-1 \times 44 \times 8$), Flatten layer ($-1 \times 352$), Dense layer 1 ($-1 \times 250$), Dense layer 2 ($-1 \times 2$), activation functions = $Relu$, epochs = 1000, batch size = 256, learning rate = 0.0001, criterion = sparse categorical crossentropy, optimizer = adam, test size = 0.3, repeat times = 30. 

\textbf{The Baseline-LSTM:} LSTM layer 1 ($-1 \times 178 \times 64$), LSTM layer 2 ($-1 \times 178 \times 32$), LSTM layer 3 ($-1 \times -1 \times 8$), Flatten layer ($-1 \times 8$), Dense layer 1 ($-1 \times 250$), Dense layer 2 ($-1 \times 2$), activation functions = $Relu$, epochs = 1000, batch size = 256, learning rate = 0.0001, criterion = sparse categorical crossentropy, optimizer = adam, test size = 0.3, repeat times = 30. 

\textbf{The Entropy-MLP:} Dense layer 1 ($4 \times 64$), Dense layer 2 ($-1 \times 64$), Dense layer 3 ($-1 \times 64$), Dense layer 4 ($-1 \times 64$), Dense layer 5 ($-1 \times 2$), activation functions = $tanh$, epochs = 2000, batch size = 256, learning rate = 0.3, criterion = Binary Cross-Entropy, optimizer = SGD, test size = 0.3, repeat times = 30.

The results of the ESRD Database are shown in Fig.~\ref{fig:ResultGADD} and Table.~\ref{tab:PerformanceGADD}. We can find that, compared with the baseline models, modeling with the entropy features can improve the recall rate, F1 score, and accuracy by up to 10.87\%, 9.74\%, and 8.41\% on average. For the model structure, compared with the Baseline-LSTM and Baseline-CNN, Entropy-MLP can reduce by 5.86 times and 1.59 times.

\begin{figure*}[t]
\vskip 0.2in
\begin{center}
\centerline{\includegraphics[width=1\textwidth]{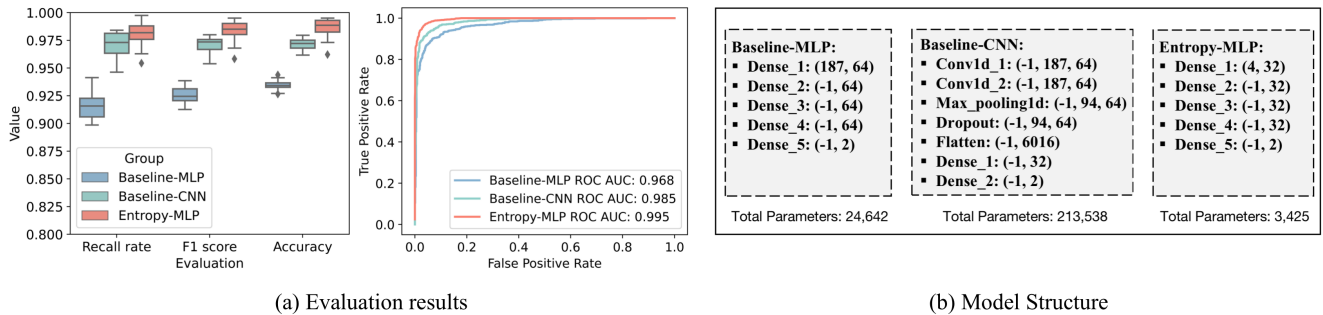}}
\caption{The results of PTBDB database. (a) is the evaluation results. (b) is the comparison of the model structure between the baseline and entropy models. From (a), we can find that the recall rate, F1 score, and accuracy of the entropy-MLP models can be improved to up to 98.08\%, 98.37\%, and 98.66\%. As we used pre-processed data with fewer data noise, the AUC-ROC performances of all the models are ideal. From (b), we can find that, the entropy-MLP model is reduced by 6.19 times and 61.35 times in terms of total parameters.}
\label{fig:ResultPTBDB}
\end{center}
\vskip -0.2in
\end{figure*}

\subsection{PTBDB}

We aim to distinguish the ordinary participants and the participants with any heart disease. The baseline models are MLP and CNN with complete data. The entropy model is MLP with PhEn, DE, ApEn, and FuzzyEn. The output of the models is the participants with any heart disease (True or False).

\textbf{The Baseline-MLP:} Dense layer 1 ($187 \times 64$), Dense layer 2 ($-1 \times 64$), Dense layer 3 ($-1 \times 64$), Dense layer 4 ($-1 \times 64$), Dense layer 5 ($-1 \times 2$), activation functions = $Relu$, epochs = 1000, batch size = 256, learning rate = 0.0001, criterion = sparse categorical crossentropy, optimizer = adam, test size = 0.3, repeat times = 30. 

\textbf{The Baseline-CNN:} Conv-1d layer 1 ($-1 \times 187 \times 64$), Conv-1d layer 2 ($-1 \times 187 \times 64$), Max pooling-1d layer ($-1 \times 94 \times 64$), Dropout layer ($-1 \times 94 \times 64$), Flatten layer ($-1 \times 6016$), Dense layer 1 ($-1 \times 32$), Dense layer 1 ($-1 \times 2$), activation functions = $Relu$, epochs = 1000, batch size = 256, learning rate = 0.0001, criterion = sparse categorical crossentropy, optimizer = adam, test size = 0.3, repeat times = 30. 

\textbf{The Entropy-MLP:} Dense layer 1 ($4 \times 64$), Dense layer 2 ($-1 \times 64$), Dense layer 3 ($-1 \times 64$), Dense layer 4 ($-1 \times 64$), Dense layer 5 ($-1 \times 2$), activation functions = $tanh$, epochs = 2000, batch size = 256, learning rate = 0.3, criterion = Binary Cross-Entropy, optimizer = SGD, test size = 0.3, repeat times = 30.

The results of the PTBDB Database are shown in Fig.~\ref{fig:ResultPTBDB} Table.~\ref{tab:PerformancePTBDB}. We can find that, compared with the Baseline-MLP and Baseline-CNN, Entropy-MLP can achieve better performance with a simpler model structure, and reduce the number of model structure parameters by 6.19 times and 61.35 times. And the Entropy-MLP can improve the recall rate, F1 score, and accuracy to 98.08\%, 98.37\%, and 98.66\%.

\begin{table}[t]
\centering
\caption{Comparison of PTBDB classification results}
\label{tab:PerformancePTBDB}
\begin{tabular}{@{}ccccc@{}}
\toprule
           &  Recall rate & F1 score & Accuracy \\ \midrule
Baseline-MLP   & 91.68±1.29\%  & 92.54±0.73\% & 93.42±0.40\%     \\
Baseline-CNN   & 97.14±1.12\%  & 97.13±0.67\% & 97.13±0.51\%      \\
Entropy-MLP   & \textbf{98.08±0.94}\%  & \textbf{98.37±0.81\%} & \textbf{98.66±0.79\%}\\\midrule
Avg Improvement   & \textbf{3.67\%}  & \textbf{3.54\%} & \textbf{3.39\%}\\\midrule
\end{tabular}
\end{table}


\section{Conclusions}\label{Conclusion} 

We propose a novel method for analyzing multivariate time-series data using information theory-based features analysis methods. Our approach utilizes entropy-based features. For applications with stochastic state transitions, we utilize Shannon's entropy of Markov chains, entropy rates of Markov chains, entropy production of Markov chains, and von Neumann entropy of Markov chains to analyze pattern changes in the data. Additionally, for applications where state transition modeling is not applicable, we used five classical entropy and entropy variants, and introduce the entropy feature selection method (by mutual information and Pearson relationship matrix).


The results show that, compared with the baseline, the entropy-based models improve the recall rate, F1 score, and accuracy on average by up to 23.01\%. We also compared the entropy-based model with state-of-the-art deep learning models on ESRD and PTBDB. And the results show that the entropy based model can achieve better performances on the recall rate, F1 score, and accuracy, with an average reduction of 18.75 times in the number of model parameters. 

The proposed pipeline offers a versatile, high-precision, and interpretable solution for analyzing time series data from the perspective of information theory, which can be applied to various forms of time series data, such as those in the fields of IoT, intelligent systems, and data security.
\section*{Acknowledgments}
This project is supported by the EPSRC PROTECT Project (grant number: EP/W031892/1), EPSRC OpenPlus Fellowship (EP/W005271/1), and the UK DRI Care Research and Technology Centre funded by MRC and Alzheimer’s Society (grant number: UKDRI-7002). The raw data from the Minder dataset was accessed using DCARTE library developed by Dr Eyal Soreq at the UK Dementia Research Institute's Care Research and Technology Centre. Yushan Huang is funded by the China Scholarship Council. Payam Barnaghi's research is also supported by the Great Ormond Street Hospital Children’s Charity Award VS0618.


%


\printbibliography

\end{document}